\documentclass{bmvc2k}

\title{Locality-Aware \hyper{} Classification}
\runninghead{Zhou \etal}{Locality-Aware \hyper{} Classification}
\addauthor{Fangqin Zhou}{f.zhou@tue.nl}{1}
\addauthor{Mert Kilickaya}{kilickayamert@gmail.com}{2}
\addauthor{Joaquin Vanschoren}{j.vanschoren@tue.nl}{3}

\addinstitution{Automated Machine Learning\\Eindhoven University of Technology\\Eindhoven, Netherlands}

\usepackage{times}
\usepackage{epsfig}
\usepackage{graphicx}
\usepackage{amsmath}

\usepackage{bbm}

\usepackage{amssymb}
\usepackage{caption}
\usepackage{slashbox}
\usepackage[dvipsnames,table]{xcolor} % loads also »colortbl«
\usepackage{tabularx,ragged2e}
\usepackage{spreadtab}
\usepackage{adjustbox}
\usepackage{booktabs}
\usepackage{multirow}
\usepackage{rotating}
\usepackage{color, colortbl}

\usepackage[inline]{enumitem}
\usepackage{color,soul}
\usepackage{glossaries}

\graphicspath{{./figures/}}

\DeclareGraphicsExtensions{.jpeg,.png,.eps,.pdf}

\newcommand{\partitle}[1]{\smallbreak\noindent\textbf{#1}}

\makeatletter
\newcommand*{\rom}[1]{\expandafter\@slowromancap\romannumeral #1@}
\makeatother

\usepackage{mathtools}

\usepackage{pifont}% http://ctan.org/pkg/pifont
\newcommand{\cmark}{\ding{51}}%
\newcommand{\xmark}{\ding{55}}%

\newcommand{\Ree}{\mathbb{R}}

\makeatletter
\newcommand*\bigcdot{\mathpalette\bigcdot@{.5}}
\newcommand*\bigcdot@[2]{\mathbin{\vcenter{\hbox{\scalebox{#2}{$\m@th#1\bullet$}}}}}
\makeatother

\newcommand{\etal}{\textit{et al}. }
\newcommand{\ie}{\textit{i}.\textit{e}., }

\usepackage{cleveref}

\crefformat{section}{\S#2#1#3} % see manual of cleveref, section 8.2.1
\crefformat{subsection}{\S#2#1#3}
\crefformat{subsubsection}{\S#2#1#3}

\usepackage{booktabs,arydshln}

\makeatletter
\def\adl@drawiv#1#2#3{%
        \hskip.5\tabcolsep
        \xleaders#3{#2.5\@tempdimb #1{1}#2.5\@tempdimb}%
                #2\z@ plus1fil minus1fil\relax
        \hskip.5\tabcolsep}
\newcommand{\cdashlinelr}[1]{%
  \noalign{\vskip\aboverulesep
           \global\let\@dashdrawstore\adl@draw
           \global\let\adl@draw\adl@drawiv}
  \cdashline{#1}
  \noalign{\global\let\adl@draw\@dashdrawstore
           \vskip\belowrulesep}}
\makeatother

\newcommand{\system}{HyLITE}
\newcommand{\hyper}{Hyperspectral}

\definecolor{lightcyan}{rgb}{0.88,1,1}
\definecolor{amber}{rgb}{1.0, 0.75, 0.0}
\definecolor{applegreen}{rgb}{0.55, 0.71, 0.0}
\definecolor{asparagus}{rgb}{0.53, 0.66, 0.42}
\definecolor{brightgreen}{rgb}{0.4, 1.0, 0.0}
\definecolor{camouflagegreen}{rgb}{0.47, 0.53, 0.42}

\definecolor{caribbeangreen}{rgb}{0.4, 1.0, 0.0}
\usepackage{amsmath}
\usepackage{bm}
\usepackage[utf8]{inputenc} % allow utf-8 input
\usepackage[T1]{fontenc}    % use 8-bit T1 fonts
\usepackage{hyperref}       % hyperlinks
\usepackage{url}            % simple URL
\usepackage{nicefrac}       % compact symbols for 1/2, etc.
\usepackage{microtype}      % microtypography
\usepackage{xcolor}         % colors

\def\1{\bm{1}}

\def\vy{{\bm{y}}}
\def\vz{{\bm{z}}}

\def\mX{{\bm{X}}}

\DeclareMathAlphabet{\mathsfit}{\encodingdefault}{\sfdefault}{m}{sl}
\SetMathAlphabet{\mathsfit}{bold}{\encodingdefault}{\sfdefault}{bx}{n}
\newcommand{\tens}[1]{\bm{\mathsfit{#1}}}

\def\tP{{\tens{P}}}

\def\tW{{\tens{W}}}

\usepackage{tikz}

\usepackage{hyperref}
\usepackage{url}
\usepackage{multirow}
\usepackage{enumitem}

\usepackage{amsthm}
\usepackage{amssymb}
\usepackage{mathtools}
\theoremstyle{plain}

\usepackage{varwidth}
\usepackage{enumitem}
\usepackage{tikz}
\usetikzlibrary{positioning}
\usetikzlibrary{shapes}
\usetikzlibrary{fit}
\usetikzlibrary{calc}

\usepackage{listings}
\usepackage{xcolor}
\usepackage{courier}

\definecolor{codegreen}{rgb}{0,0.6,0}
\definecolor{codegray}{rgb}{0.5,0.5,0.5}
\definecolor{codeblack}{rgb}{0.,0.,0.}
\definecolor{codepurple}{rgb}{0.58,0,0.82}
\definecolor{backcolour}{rgb}{0.95,0.95,0.92}

\lstdefinestyle{mystyle}{
    backgroundcolor=\color{backcolour},   
    commentstyle=\color{codegreen},
    keywordstyle=\color{codeblack},
    numberstyle=\tiny\color{codegray},
    stringstyle=\color{codepurple},
    basicstyle=\ttfamily\footnotesize,
    breakatwhitespace=false,         
    breaklines=true,                 
    captionpos=b,                    
    keepspaces=true,                 
    numbers=left,                    
    numbersep=5pt,                  
    showspaces=false,                
    showstringspaces=false,
    showtabs=false,                  
    tabsize=2,
    aboveskip=0pt,
    belowskip=-3pt
}

\newcommand{\Quote}[2]{
}

\usepackage{pifont}% http://ctan.org/pkg/pifont
\begin{document}
\maketitle
\begin{abstract}

\hyper{} image classification is gaining popularity for high-precision vision tasks in remote sensing, thanks to their ability to capture visual information available in a wide continuum of spectra. Researchers have been working on automating \hyper{} image classification, with recent efforts leveraging Vision-Transformers. However, most research models only spectra information and lacks attention to the locality (\ie neighboring pixels), which may be not sufficiently discriminative, resulting in performance limitations. To address this, we present three contributions: \textit{i)} We introduce the \textbf{Hy}perspectral \textbf{L}ocality-aware \textbf{I}mage \textbf{T}ransform\textbf{E}r (\system{}), a vision transformer that models both local and spectral information, \textit{ii)} A novel regularization function that promotes the integration of local-to-global information, and \textit{iii)} Our proposed approach outperforms competing baselines by a significant margin, achieving up to $10\%$ gains in accuracy. The trained models and the code are available at \href{https://github.com/zhoufangqin/HyLITE.git}{HyLITE}. 
% \footnote{This paper is accepted by BMVC2023}
% All the models, code, and data will be made public upon publication.

\vspace{-3mm}

\end{abstract}

\section{Introduction}

\hyper{} Imaging (HSI) is capable of remotely capturing a large field of view by sampling the continuum of the electromagnetic spectrum. As a result, HSI provides fine-grained information that is not typically available in conventional RGB images. This ability to leverage such fine-grained information has led to breakthroughs in various industries, including monitoring plants in agriculture \citep{dale2013hyperspectral, xie2015detection, furbank2021wheat, wang2019early, nagasubramanian2019plant, wang2019early}, remote sensing of the Earth's surface \citep{goetz2009three, applicationremotesensing}, and better navigation and vision in robotics \citep{jakubczyk2022hyperspectral, trierscheid2008hyperspectral}.

%to increase the sustainability. For example, practitioners use HSI to monitor the status of plants~\citep{ferwerda2005charting,furbank2021wheat}, to diagnose leaf diseases~\citep{furbank2021wheat}, or to categorize waste to automate recycling~\citep{ bonifazi2019hyperspectral}. 

\begin{figure}[t]
    \centering
\includegraphics[width=\textwidth]{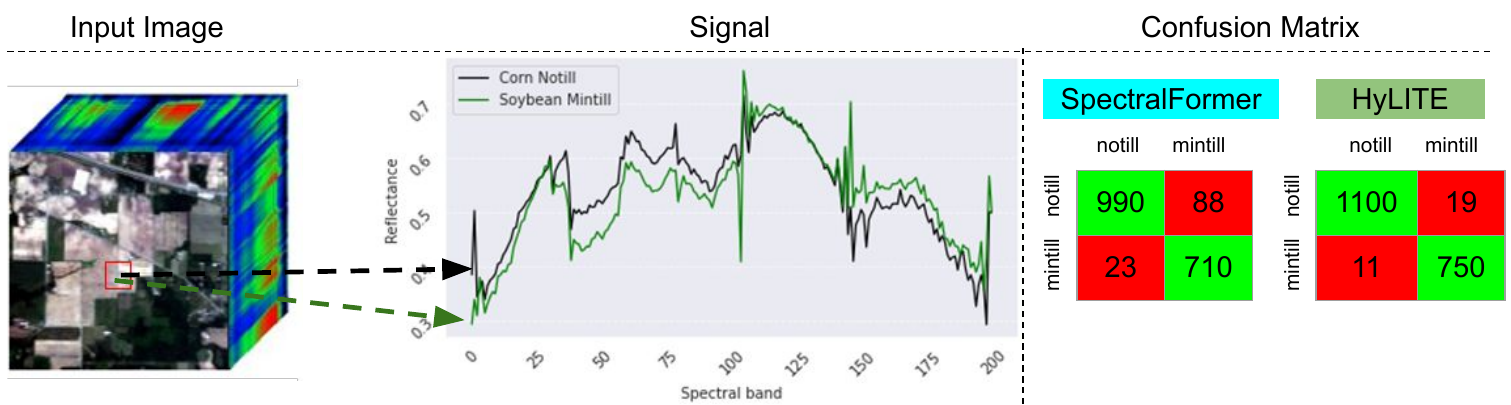}
    %\caption{The goal of this paper is \hyper{} image classification. (a) Given an input on the top-left, the goal is to identify the category per-pixel, such as $Corn$, $Corn Notill$ and $Soybean Mintill$. (b) Recent research relies on modeling spectral relations for classification. In this paper, we advocate that spectral information alone is not sufficient, as two pixels with distinct categories may exhibit highly similar spectral signals. To that end, in this paper, we propose, \system{}, that improves the locality-awareness of the \hyper{} image transformers.}
    \caption{(\textbf{Left}) A \hyper{} image cube, with the spectral signals of two pixels from separate categories (`notill' and `mintill') from the Indian Pines dataset~\citep{hyperdata}. Observe how their spectral signature is highly similar. (\textbf{Right}) Confusion matrices of SpectralFormer~\citep{hong2021spectralformer} and \system{} (Ours). SpectralFormer confuses these two classes, as it solely relies on the spectral signal. \system{} disambiguates the classes by incorporating locality information.}
    %\caption{(\textbf{Left}) A \hyper{} image cube, with the spectral signals of two highly similar categories (`notill' and `mintill')~\citep{hyperdata}.  (\textbf{Right}) SpectralFormer~\citep{hong2021spectralformer} confuses these two classes, whereas our \system{} disambiguates the confusion by incorporating locality.}
    \label{fig:teaser}
\end{figure}

The first deep HSI-based techniques adopted Convolutional Neural Networks (CNNs) to learn representations, either in a discriminative \citep{zhang2016deep,li2017spectral, zhong2017spectral, nagasubramanian2019plant,zhang2017spectral, paoletti2018deep, zhao2016spectral} or generative manner \citep{wang2019early, ding2021improving}. However, the performance of CNNs has been limited, due to the limited receptive field of CNNs \citep{ViT, aleissaee2023transformers}, which cannot model long-range dependencies across spatial-spectral dimensions. To model long-range dependencies, self-attention is incorporated into CNNs \citep{mei2019spectral, sun2019spectral, zhu2020residual, xie2021multilayer, zheng2022rotation}, significantly increasing the receptive field of CNNs by enabling any pixel or spectrum to aggregate information from any other pixel or spectrum within the input cube. However, the receptive field of the backbone CNN representation is still too limited for HSI.

To overcome this limitation, most recent techniques rely solely on the Vision-Transformer \citep{ViT}, which is a stack of multi-head self-attention modules. One notable example is SpectralFormer, which models the interactions across spectral bands to classify an input pixel \citep{hong2021spectralformer}. However, modeling spectral interactions alone can be insufficient for classification when pixels with different classes have very similar spectral signatures.
%In Figure~\ref{fig:teaser}, we plot the spectral signal of two pixels from Indian Pines dataset~\citep{hyperdata}, from notill and mintill classes. These pixels exhibit highly similar spectral signature, which confuses the strong baseline of SpectralFormer~\citep{hong2021spectralformer}. However, a \hyper{} signal offers much more than spectral signal: The local information around the pixel (\ie the representation of the nearby pixel) may help in disambiguating the target category. Also, the global information within the \hyper{} signal could be useful (\ie the set of potential categories present within the input image). More importantly, modeling the relationship between local and spectral information is important, as not both may be concurrently discriminative for a given pixel.

For instance, in Figure~\ref{fig:teaser}, we show the spectral signal of two pixels from the Indian Pines dataset~\citep{hyperdata}, belonging to the `notill' and `mintill' classes, respectively. These pixels exhibit highly similar spectral signatures, which confuses the SpectralFormer~\citep{hong2021spectralformer}. Fortunately, besides the spectral signal, \hyper{} images also provide the local information around a pixel, such as the representation of nearby pixels, which may help in disambiguating the target categories. Additionally, the global information within the \hyper{} signal, such as the set of potential categories present within the input image, could be useful. Hence, modeling the relationship between local and spectral information can help discriminate between the possible classes for a given pixel.
 
%In our case, we are able to disambiguate the confusion by incorporating local (\ie representation of nearby pixels) as well as global (\ie the set of categories present within the input patch) context.  

%\partitle{Existing Methods: Limitation of Locality.}
%Both local and global information play important roles in image classification tasks. As well studied, CNNs are good at extracting local features via convolution kernel tricks but lack global information reasoning. On the other hand, transformer models succeed in maintaining global connectivity via self-attention over the entire input sequence but usually missing local information. Recently, many ideas have been proposed to bring locality to ViT ~\citep{li2021localvit, chu2021twins, chen2021regionvit, yao2022dual}, and they have shown that incorporating locality into ViT leads to a significant performance gain. However, combining locality with modern transformer models is still under studied in HSI classification. Thus, we propose our approach: Locality-Aware Transformers for HSI Classification (LAT-HSIC).

%\partitle{\system{}} 

We propose a \textbf{Hy}perspectral \textbf{L}ocality-aware \textbf{I}mage \textbf{T}ransform\textbf{E}r (\system{}), which extends SpectralFormer in three major ways. Firstly, \system{} models the relationships between local (spatial) and spectral representations, so that when spectral information is insufficient, local information can come into play. Secondly, we model the relationships between \textit{local} and \textit{global} representations with a novel regularization objective. Our regularization loss enforces local and distant pixels and spectra to aggregate information from each other, effectively improving representational capacity. Finally, we extensively evaluate our method on three standard HSI benchmarks and show that \system{} establishes a new State-of-the-Art in HSI classification. 

In summary, this paper makes three main contributions: 

\begin{enumerate}[label=\Roman*.]

\item We propose \textbf{Hy}perspectral \textbf{L}ocality-aware \textbf{I}mage \textbf{T}ransform\textbf{E}r (\system{}), a novel architecture that can model the local-spectral relationships in \hyper{} data.  
\item We equip \system{} with a novel local-global regularization objective, to balance global and local spectral information.  
\item We conduct experiments on three well-established benchmarks, and show that \system{} significantly improves over the competitive SpectralFormer baseline, across \textit{all} benchmarks and metrics. For example, we improve the overall accuracy by $10.83\%$ on Indian Pines~\citep{hyperdata}, by $3.41\%$ on Houston2013~\citep{houston2013}, and by $6.64\%$ on Pavia~\citep{hyperdata}. 

\end{enumerate}

%\mert{Expand on the importance of locality, difference of \hyper{} to rgb, the details of the contribution and empirical benefits.}

%  \begin{figure}[t]
%     \centering
% \includegraphics[width=0.5\textwidth]{figures/dummy.png}
% \caption{Method motivation.}
% \end{figure}

\section{Related Work}

\partitle{Vision Transformers.} Vision-Transformers (ViTs) are a direct translation of language transformers in NLP~\citep{nlp, devlin2018bert} to computer vision~\citep{ViT}. The key building block of ViT is the self-attention layer~\citep{vaswani2017attention}, which consists of three subnetworks, namely Query, Key and Value networks. Query and Key networks compute the attention across the input signal, which is then used to modulate the input signal using the Value network. Since the release of ViT, many variants have been proposed~\citep{DeiT, liu2021swin, iGPT, yuan2021tokens, bao2021beit, PVT}. However, the majority of these are limited to processing RGB images. In this work, our model input consists of pre-processed \hyper{} image patches, which have a much lower spatial resolution (\ie $7\times7$ \textit{vs.} $512\times512$) and much higher spectral resolution (\ie $200$ \textit{vs.} $3$). These fundamental differences demand specialized architectures and loss functions. To that end, in this work, we develop new components combined with a novel loss function to learn the relationships between local and spectral information.

\partitle{\hyper{} Image Transformers.} ViTs have previously been used in state-of-the-art \hyper{} image classification architectures~\citep{hong2021spectralformer,sun2022spectral,ibanez2022masked}. Hong~\etal~\citep{hong2021spectralformer} propose SpectralFormer, which treats each spectrum as a distinct token, and models spectrum-to-spectrum attention. However, HSI offers much more than spectrum-only information: it also contains spatial information, such as the local neighborhood of a pixel. To that end, Sun~\etal~\citep{sun2022spectral} augment SpectralFormer with spatial attention. However, the mere combination of spectral and spatial attention is suboptimal. As we will demonstrate, regularizing local pixel representations with respect to global information matters greatly. Finally, Ibanez~\etal~\citep{ibanez2022masked} propose MAEST, which pre-trains a transformer backbone by predicting masked wavebands with Masked-AutoEncoders~\citep{he2022masked} prior to labeled fine-tuning. Such pre-training promotes locality and improves performance. However, self-supervised pre-training is computationally costly~\citep{singh2023effectiveness}. Our \system{} model combines spectral and spatial attention while also introducing local-global regularization, which achieves better performance without requiring pre-training.

%%%% V4 %%%%%%%%%%%%%%%%%%%%%%%%%%%%%%%%%%%

%\input{3-method-v1}
%\vspace{-1mm}

\section{Hyperspectral Locality-aware Image TransformEr}
\label{sec:method}

\begin{figure}[h]
    \centering
    \includegraphics[width=0.95\textwidth]{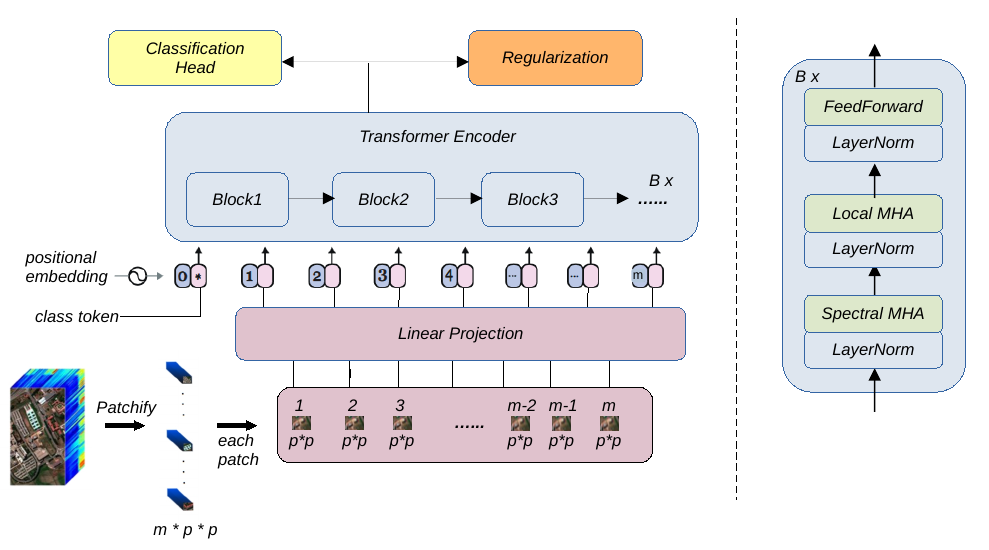}
\caption{An overview of \system{}. \textcolor{magenta}{\textit{i). Preprocessing}}: The input image is patchified, linearly projected, and appended with a classifier token and a positional embedding. \textcolor{cyan}{\textit{ii). Representation}}: The input is processed by identical spectral and local multi-head attention (MHA) blocks. \textcolor{amber}{\textit{iii). Classification}}: At the end, the representation of the classifier token is mapped to a distinct category, such as $\{grass, road\}$. \textcolor{orange}{\textit{iv). Regularization}}: To further promote locality, we apply our novel regularization on top of the learned token representations.}
    \label{fig:architecture}
\end{figure}

% Mert: Major update to the notation is necessary: 
% Dont start from I, start directly from x. 
% Follow matrix, tensor, vector notation directly from preamble.tex
%%%%%%%%%%%% V1 %%%%%%%%%%%%%%%%%%%%%%%%%%%%%%%%%%%%%%%%%%%%%%%%%%%%%%%%%%%%%%%

%An overview of \system{} is given in Figure~\ref{fig:architecture}. We strictly follow the protocol in~\citep{hong2021spectralformer}, which treats HSI as an image-level classification task. Suppose we are given an image-label pair $(\mX, \vy)$. Here the input is a low-resolution square image $\mX \in\Ree^{p\times p\times m}$ where $p$ is the spatial resolution (\ie $7\times 7$) and $m$ is the spectral resolution (\ie $200$). There are $\text{N}$ such images, as they are sampled from a large-resolution \hyper{} image by overlapping patches. Each image is labeled by the category of its center pixel $\vy\in\Ree^{c\times1}$ from $c$ potential classes, such as grass, road, etc. The goal is to train a vision transformer $f_{\theta}$ with parameters $\theta$ to predict the image label $\vy^{\prime} = f_{\theta}(\mX)$, where $\vy^{\prime}$ denotes the predictions. We detail the flow of the model below. 

An overview of \system{} is given in Figure~\ref{fig:architecture}. We strictly follow the protocol in~\citep{hong2021spectralformer}, treating HSI as an image-level classification task. Given an image-label pair $(\mX \in\Ree^{p\times p\times m}, \vy \in\Ree^{c})$ where input $\mX$ is a low-resolution square image with spatial resolution $p\times p$ (\ie $7\times 7$) and spectral resolution $m$ (\ie $200$), sampled from a high-resolution \hyper{} image by overlapping patchifying. Each patch is labeled by the category of its center pixel from $c$ potential classes, such as grass, road, etc. The goal is to train a vision transformer $f_{\theta}$ with parameters $\theta$ to predict the image label $\vy^{\prime} = f_{\theta}(\mX)$, where $\vy^{\prime} \in\Ree^{c}$ denotes the predictions. We train our network in four main steps, which are detailed below.  

\partitle{1) \textul{Preprocessing}.} We first transform the input image by transposing and flattening it in the spatial dimension: $\mX \in\Ree^{m\times p^{2}}$, thus yielding $m$ spectral tokens with dimensionality $p^{2}$: $\mX = [\vz^{1}; \vz^{2}; ...; \vz^{m}]$ where $\vz^{i}\in\Ree^{1\times p^{2}}$. We use $\mX^{i}_{j}$ to denote the spectral token $\vz^{i}$ at the $j$th transformer block. Next, we embed the tokens in $\Ree^{d}$ using a linear projection $\tW\in\Ree^{p^{2}\times d}$, insert a learnable global classifier token $\vz^{0} \in\Ree^{1\times d}$, and add a learnable position tensor $\tP\in\Ree^{(m+1)\times d}$ to each token. As such, the first transformer block takes the form: 
\begin{align}
    \mX_{0} = [\vz^{0}; \vz^{1}\tW; ..., \vz^{m}\tW] + \tP
\end{align}

%\noindent where $\tW\in\Ree^{p^{2}\times d}$ is the linear-projection matrix, $\vz^{0} \in\Ree^{1\times d}$ is the global classifier token, and $\tP\in\Ree^{(m+1)\times d}$ is the learnable position tensor. %After preprocessing, the input is ready to be processed by representation blocks. 

\partitle{2) \textul{Representation}.} Each of the $B$ transformer blocks consists of three subsequent layers, namely Spectral-Attention $\text{S}(\cdot)$, Local-Attention $\text{L}(\cdot)$ and Feed-Forward $\text{F}(\cdot)$ layers, applied one after the other: $\mX_{b} = \text{F}(\text{L}(\text{S}(\mX_{b-1}))) \in\Ree^{(m+1)\times d}$ where  $b=1\ldots B$. Before each layer, a LayerNorm~\citep{ba2016layer} is applied, which we omit for clarity. Below, we detail each layer. 

% softmax(\frac{qk^{T}}{\sqrt{d}})v$
First, the spectral Multi-Head Attention (MHA) layer combines information from across the \textit{spectral} dimension. It is implemented via self-attention. Formally: 
\begin{align}
    \text{S}(\mX) = \text{softmax}(\frac{(\mX\tW_{q}^{s}) (\mX\tW_{k}^{s})^{T}}{\sqrt{d}})(\mX\tW_{v}^{s})
\end{align}
\noindent where $\{\tW_{q}^{s}, \tW_{k}^{s}, \tW_{v}^{s}\}$ are the linear query-key-value projections, respectively, all with tensor dimensionality $\Ree^{d\times d}$, where $\sqrt{d}$ is a scaling factor, and $\text{softmax}(\cdot)$ is the softmax operator.% within the last dimension. 

Second, the local MHA layer combines information across the \textit{local} (spatial) dimension. It is also implemented via self-attention. Formally: 
\begin{align}
    \text{L}(\mX) = \left(\text{softmax}(\frac{(\mX^{T}\tW_{q}^{l})(\mX^{T}\tW_{k}^{l})^{T}}{\sqrt{d}})(\mX^{T}\tW_{v}^{l})\right)^{T}
\end{align}
\noindent where $\{\tW_{q}^{l}, \tW_{k}^{l}, \tW_{v}^{l}\}$ are again linear query-key-value projections, with tensor dimensionality $\Ree^{(m+1)\times (m+1)}$. 

Finally, the feed-forward layer consists of two linear layers and generates the output of each block with dimensionality $\Ree^{(m+1)\times d}$. After the last block, the token representations are passed to both a classification and regularization head.

\partitle{3) \textul{Classification}.} We generate predictions by learning a linear mapping from the global token representation of the last layer to the classification labels: $\vy^{\prime} = \mX^{0}_{B}C$, where $C\in\Ree^{d\times c}$ is the classifier head. Using the predictions and the final representations, the model is trained by minimizing the following objective: 
\begin{align}
    \mathcal{O} = \text{CE}(\vy, \vy{\prime}) + \lambda\cdot\text{Reg}(\mX_{B})
\end{align}
\noindent where $\text{CE}(\cdot)$ is the standard cross-entropy loss, $\text{Reg}(\cdot)$ is our novel local-global regularization objective, and $\lambda$ attenuates the regularization strength. In the remainder of this paper, we set $\lambda=1$. The effects of different $\lambda$ values are reported in the supplemental materials. 

%\partitle{4) \textul{Regularization}.} \hyper{} classification is different from traditional image classification, in the sense that spatial input resolution is quite small (\ie $7\times 7$) whereas spectral resolution is large (\ie $200$). To that end, when training a vision-transformer, we observe the following phenomenon: The global classifier token attends to only a few spectra, neglecting the local context. In meanwhile, spectral tokens mostly attend to their self-representation, neglecting  the global context. To enforce aggregation of local-global context, we propose the following simple objective: 

\partitle{4) \textul{Regularization}.} During the training of \system{}, we observed that the \textit{spectral} tokens mostly attend to their \textit{self}-representation, hence focusing on the local context and ignoring the global context. Moreover, the \textit{global} token mostly attends to a few specific tokens, ignoring the rest of the local context. This indicates that the spectral and global token representations diverge throughout the blocks, and hence we name this phenomenon "attentional divergence". To prevent such divergence and improve performance, we propose the following regularization objective: 
% \begin{align}
%     \text{Reg}(\mX_{B}) = \frac{1}{m}\sum_{i=1}^m\left\| \mX_{B}^{0} - \mX_{B}^{i} \right\|_{2}^{2}
% \end{align}
% \noindent To satisfy this loss function, first, the global output token $\mX_{B}^{0}$ has to be similar to \textit{all} of the $m$ spectral output tokens $\mX_{B}^{i>0}$. This forces the global token to aggregate information from a diverse set of tokens, effectively incorporating the locality. Simultaneously, the spectral tokens have to be similar to the global token. This forces the spectral tokens to aggregate information from the \textit{other} spectral tokens, effectively incorporating the globality. As we will demonstrate, this regularization term has a significant effect on classification performance.
\begin{align}
    \text{Reg}(\mX_{B}) = \left\| \mX_{B}^{0} - \frac{1}{m}\sum_{i=1}^m\mX_{B}^{i} \right\|_{2}^{2}
\end{align}
\noindent To minimize this loss function, the global output token $\mX_{B}^{0}$ should be close to the center of the spectral tokens. Hence, the gradients will nudge the representations of the global and spectral tokens closer together, causing them to converge rather than diverge, and aggregating information from each other, thus incorporating globality in the learning process.
%This prevents that the global token attends solely to a few spectral tokens in the final stage. Simultaneously, the local tokens should also be close to the global token, which forces the spectral tokens to aggregate information from the global token, effectively incorporating the globality. 
Notably, the use of a learned global token works differently (and performs much better) than simply applying average pooling over all spectral tokens. It better balances acquired global and local knowledge, while average pooling might cause information loss. As we will demonstrate, this approach has a significant positive effect on classification performance.

\section{Experimental Setup}

%\mert{Shorten this part significantly and move most of the details to supplementary.}

We evaluate our method on multiple datasets, with multiple metrics, and against multiple strong baselines. The supplementary provides further details on implementation and setup. 

\partitle{$\triangleright$Datasets:} We evaluate our model on three well-established HSI datasets. \textit{i). Indian Pines~\citep{hyperdata}:} consists of $224$ spectral bands with $145\times145$ spatial resolution. It includes $16$ classes, $695$ training, and $9k$ testing samples. \textit{ii). Houston2013~\citep{houston2013}:} consists of $144$ spectral bands with $349\times1905$ spatial resolution. It includes $15$ classes, $2k$ training, and $12k$ testing samples.  \textit{iii). Pavia University~\citep{hyperdata}:} consists of $103$ spectral bands with $610\times340$ spatial resolution. It includes $15$ classes, $3k$ training, and $40k$ testing samples. 

\partitle{$\triangleright$Metrics: } We rely on the standard metrics, namely Overall Accuracy (OA), Average Accuracy (AA) and Kappa Coefficient ($k$). OA denotes the total number of correctly predicted samples over all samples, while AA denotes the average accuracy of each class. 

\partitle{$\triangleright$Baselines:} We mainly compare \system{} against the state-of-the-art methods SpectralFormer~\citep{hong2021spectralformer} and $MAEST^{*}$~\citep{ibanez2022masked}. We also compare against conventional methods (\ie kNN, RF), and CNN-based techniques (N-D CNNs) for the sake of completeness.

\section{\hyper{} Image Classification}

\subsection{Comparison to the State-of-the-Art}

%We have four observations: \textit{i).} \system{} obtains the State-of-the-Art result, by a large margin, across all three datasets, across all three metrics. \textit{ii).} The performance gap against the most competitive baseline, SpectralFormer is especially remarkable, as we improve between $3-10\%$ in overall accuracy in respective benchmarks. This confirms our hypothesis that indeed SpectralFormer is missing the crucial information present within the locality.  \textit{iii).} The improvement of \system{} is even more pronounced on Indian Pines, with $10.83\%$ and $9.30\%$ in overall and average accuracies respectively. This indicates that \system{} is able to leverage limited amount of training data, as amongst all three benchmarks, Indian Pines has the smallest amount of training data (\ie only $693$ instances). \textit{iv).} Finally, we observe that \system{} even outperforms MAEST~\citep{ibanez2022masked}, which performs a computationally expensive self-supervised pre-training prior to fine-tuning. This indicates with once discriminative inductive biases are present within the model, the need for expensive self-supervised pre-training is mitigated. 

Firstly, we compare our method against several baselines from the HSI literature in Table~\ref{tab:sota1}. We use the exact same evaluation procedure as used in~\citep{hong2021spectralformer} to get a fair comparison.
We can make four observations based on these results: \textit{i).} \system{} achieves State-of-the-Art performance by a significant margin across all three datasets and all three metrics. \textit{ii).} Our performance surpasses that of the most competitive baseline, SpectralFormer, by $3-10\%$ in overall accuracy in the respective benchmarks, confirming our hypothesis that SpectralFormer lacks crucial locality information. \textit{iii).} Third, the improvement of \system{} is particularly noteworthy on the Indian Pines dataset, with Overall Accuracy and Average Accuracy improving by $10.83\%$ and $9.30\%$, respectively. This suggests that adding locality information helps to learn more effectively from small amounts of training data, since Indian Pines has the smallest amount of training data among the three benchmarks, with only $695$ instances. We will further explore this in the next section. \textit{iv).} \system{} even outperforms $MAEST$~\citep{ibanez2022masked}, which performs computationally expensive self-supervised pre-training prior to fine-tuning. This indicates that once discriminative inductive biases are present within the model, self-supervised pre-training is no longer needed. 

We conclude that incorporating locality is highly valuable for \hyper{} image classification, as evidenced by the significant improvement over competitive baselines across all benchmarks.

%%%%%%%%%%%%% V1 %%%%%%%%%%%%%%%%%%%%%%%%%%%%%%%%%%%%%

\begin{table}[t]
\centering
\resizebox{\columnwidth}{!}{
\begin{tabular}{llllllllll}
\toprule

 & \multicolumn{3}{c}{Indian Pines} & \multicolumn{3}{c}{Houston2013} & \multicolumn{3}{c}{Pavia University} \\ 
 \cmidrule(lr){2-10} 

 & OA & AA & Kappa & OA & AA & Kappa & OA & AA & Kappa \\ 
 \cmidrule(lr){2-4} \cmidrule(lr){5-7}  \cmidrule(lr){8-10}

kNN & 59.17  & 63.90  & 0.54  & 77.30  & 78.28  & 0.75  & 70.53  & 79.68  & 0.62 \\ 
RF & 69.80 & 76.78 & 0.65 & 77.48 & 80.35 & 0.75 & 69.67 & 80.18 & 0.62 \\
SVM & 72.36 & 83.16 & 0.68 & 76.91 & 78.99 & 0.79 & 70.82 & 84.44 & 0.64 \\ 
\cdashlinelr{1-10}
1-D CNN & 70.43 & 79.60 & 0.66 & 80.04 & 82.74 & 0.78 & 75.50 & 86.26 & 0.69 \\ 
2-D CNN  & 75.89 & 86.64 & 0.72 & 83.72 & 84.35 & 0.82 & 86.05 & 88.99 & 0.81 \\ 
RNN  & 70.66 & 76.37 & 0.66 & 82.23 & 85.04 & 0.81 & 77.13 & 84.29 & 0.71 \\ 
miniGCN & 75.11 & 78.03 & 0.71 & 81.71 & 83.09 & 0.80 & 79.79 & 85.07 & 0.73 \\ 
\cdashlinelr{1-10}

ViT & 71.86 & 78.97 & 0.68 & 80.41 & 82.50 & 0.78 & 76.99 & 80.22 & 0.70 \\ 
SpectralFormer & 78.97 & 85.39 & 0.76 & 85.08 & 86.39 & 0.83 & 84.64 & 86.75 & 0.79 \\ 
\rowcolor{lightcyan} \system{} (Ours) 
& $89.80$   
& $94.69$  
& $0.88$  
& $88.49$  
& $89.74$  
& $0.87$  
& $91.28$  
& $92.25$  
& $0.88$ \\ 

 $\Delta$  & \textcolor{red}{$10.83$} & \textcolor{red}{$9.30$} & \textcolor{red}{$0.12$}   
& \textcolor{red}{$3.41$} & \textcolor{red}{$3.35$} & \textcolor{red}{$0.03$} 
& \textcolor{red}{$6.64$} & \textcolor{red}{$5.50$} & \textcolor{red}{$0.08$}  \\ 

\cdashlinelr{1-10}
$MAEST$  & 82.12  & 87.63  & 0.79  & 83.61  & 84.89  & 0.82  & 87.20  & 89.91  & 0.83 \\

\bottomrule
\end{tabular}}

%\caption{Comparison to the state-of-the-art models on three HSI datasets. Our model leads to significant improvement across all datasets across all metrics, confirming the benefit of incorporating locality. }

%\caption{Comparison to the State-of-the-Art. We provide the results from all baselines, as well as the performance gap ($\Delta$) with SpectralFormer. Our model outperforms all techniques by a large-margin regardless of the dataset and metric. The improvement is even more pronounced when the training data is smaller (only $693$ training instances in Indian Pines) or the testing data is much larger ($40$k testing instances in Pavia University). Surprisingly, our model even outperforms $MAEST^{*}$~\citep{ibanez2022masked}, which performs computatinally expensive self-supervised pre-training prior to fine-tuning. This confirms our hypothesis that locality matters greatly in HyperSpectral image classification.}

\caption{Comparison against the State-of-the-Art. We provide the results from all baselines, as well as the performance gap ($\Delta$) with SpectralFormer. Our model consistently outperforms all techniques by a wide margin across all datasets and evaluation metrics. Notably, our model most significantly outperforms SpectralFormer when the training set is small (e.g., only $695$ training instances in Indian Pines). % or larger testing sets (e.g., $40k$ testing instances in Pavia University). 
Surprisingly, our model even outperforms the computationally much more expensive self-supervised pre-training approach of $MAEST$. These findings support our hypothesis that spatial locality plays a crucial role in \hyper{} image classification.}

\label{tab:sota1}
\end{table}

%%%%%%%%%%%%% V2 %%%%%%%%%%%%%%%%%%%%%%%%%%%%%%%%%%%%%

\subsection{Evaluation of Sample Efficiency}

\begin{figure}[t]
    \centering
\includegraphics[width=0.75\textwidth]{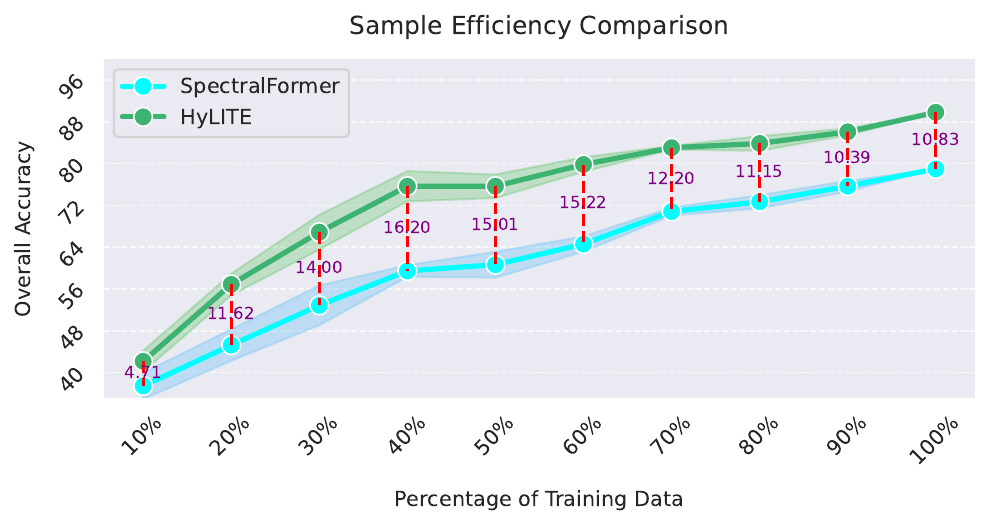}
\caption{Comparing the sample efficiency of \system{} and SpectralFormer~\citep{hong2021spectralformer} on Indian Pines~\citep{hyperdata}. \system{} outperforms across all subsets, confirming the efficacy of locality for learning from limited examples.}
\label{fig:ip_sub}
\end{figure}

In the previous section, we focused on learning on all available data. However, in \hyper{} imaging, sample efficiency matters greatly, as data collection and annotation are costly. To that end, we compare the performance of \system{} with SpectralFormer on Indian Pines, by varying the size of the training set within $\{10\%, 20\%, ..., 100\%\}$. We repeat the experiment $4$ times with different random training samples, and present the mean and standard deviation in Figure~\ref{fig:ip_sub}. 

%Secondly, we compare the sample efficiency of our model to SpectralFormer~\citep{hong2021spectralformer}. We vary the size of the training dataset between $\{10\%, 20\%, ..., 100\%\}$ on Indian Pines~\citep{xxx}. Results are presented in Figure~\ref{fig:ip_sub}. 

From this, we can make two observations: \textit{i).} The performance of both models degrades drastically as the size of the training set shrinks (\ie $10\%$). This is expected, as Vision-Transformers require a sufficient number of exemplars to generalize. \textit{ii).} \system{} is always superior to SpectralFormer across all subsets. This indicates that incorporating locality is useful in improving the sample efficiency of \hyper{} image classifiers. 

Hence, we conclude that incorporating locality not only improves accuracy, but also improved sample efficiency of \hyper{} imaging, which is promising for low-shot learning applications.

%As can be seen \mert{Comment on the results.}
%\subsection{Further Analysis}
%This section analyses the remarkable performance of \system{}. 

\subsection{Ablation Analysis}
%Thirdly, we ablate different components in our model. Results are presented in Table~\ref{tab:ablation}. 

We ablate the position and the components of \system{} in Table~\ref{tab:ablation}. 

%Previous sections highlight the remarkable performance of \system{}. To understand the relative contribution of different components, here we provide our ablation study in Table~\ref{tab:ablation}. 

\begin{table}[htbp]
\centering
\resizebox{\columnwidth}{!}{
\begin{tabular}{llllllllll}
\toprule

 & \multicolumn{3}{c}{Indian Pines} & \multicolumn{3}{c}{Houston2013} & \multicolumn{3}{c}{Pavia University} \\ 
 \cmidrule(lr){2-10} 

 & OA & AA & Kappa & OA & AA & Kappa & OA & AA & Kappa \\ 
 \cmidrule(lr){2-4} \cmidrule(lr){5-7}  \cmidrule(lr){8-10}

%\multicolumn{10}{c}{\textit{Positional Embedding}} \\
\textit{Positional Embedding} \\ 
\cdashlinelr{1-10}
No  Embedding & 79.63  & 85.35  & 0.77  & 84.09  & 85.96  & 0.83  & 84.69  & 87.18  & 0.79 \\ 
Fixed Embedding & 85.30  & 88.56  & 0.83  & 83.46  & 85.26  & 0.82  & 87.36  & 85.64  & 0.83 \\ 
\rowcolor{lightcyan} Learned Embedding & 89.80  & 94.69  & 0.88  & 88.49  & 89.74  & 0.88  & 91.28  & 92.25  & 0.88 \\ 
\midrule 
%\multicolumn{10}{c}{\textit{Components}} \\ 
\textit{Components} \\ 
\cdashlinelr{1-10}
local-att(\xmark) \& local-reg(\xmark) & 78.97 & 85.39 & 0.76 & 85.08 & 86.39 & 0.83 & 84.64 & 86.75 & 0.79 \\ 
local-att(\cmark) \& local-reg(\xmark) & 85.37  & 90.09  & 0.83  & 87.69  & 89.18  & 0.87  & 87.78  & 91.06  & 0.84 \\ 
local-att(\xmark) \& local-reg(\cmark) & 83.00  & 89.40  & 0.81  & 87.13  & 88.33  & 0.86  & 87.76  & 91.72  & 0.84 \\ 
\rowcolor{lightcyan} local-att(\cmark) \& local-reg(\cmark) & 89.80  & 94.69  & 0.88  & 88.49  & 89.74  & 0.88  & 91.28  & 92.25  & 0.88 \\ 

\bottomrule
\end{tabular}}

\caption{Ablation Study. Learning the positional embedding, as well as combining our local attention (local-att) with local-global regularization (local-reg) matters.}
\label{tab:ablation}
\end{table}

\partitle{Positional Embedding.} Here, we try to understand the contribution of positional information in \system{}. First, removing positional information leads to a drastic performance drop, as expected. This indicates \system{} greatly utilizes location information. Secondly, learning positional information yields much better performance than fixing positional information. This indicates \system{} learns to adjust the relative contribution of each token within the input. 

%As the sole purpose of \system{} is to incorporate locality, an important aspect of the model is to efficiently leverage positional information. To understand this, we run the results without positional embedding, with fixed positional embedding, and finally learned positional embedding. Removing positional embedding leads to a significant drop in performance, as is the case with any vision transformers. We also observe that learning the positional embedding matters, which indicates the model learns to adjust the contribution of each position accordingly. This confirms that to effectively incorporate locality, position information is crucial. 

\partitle{Components.} We ablate the local attention (local-att) and regularization (local-reg) components to understand their relative contribution. Firstly, including either local attention or local-global regularization helps greatly. However, incorporating both leads to a significant gain, indicating that regularizing the local-attention representation matters in \hyper{} image classification.  

We conclude that \system{} is able to incorporate positional information efficiently, and regularizing the local-attentional representation is critical for classification accuracy. 

\subsection{Category-level Analysis}
%\partitle{Category-level Comparison.} 

\begin{figure}[h]
    \centering
\includegraphics[width=0.6\textwidth]{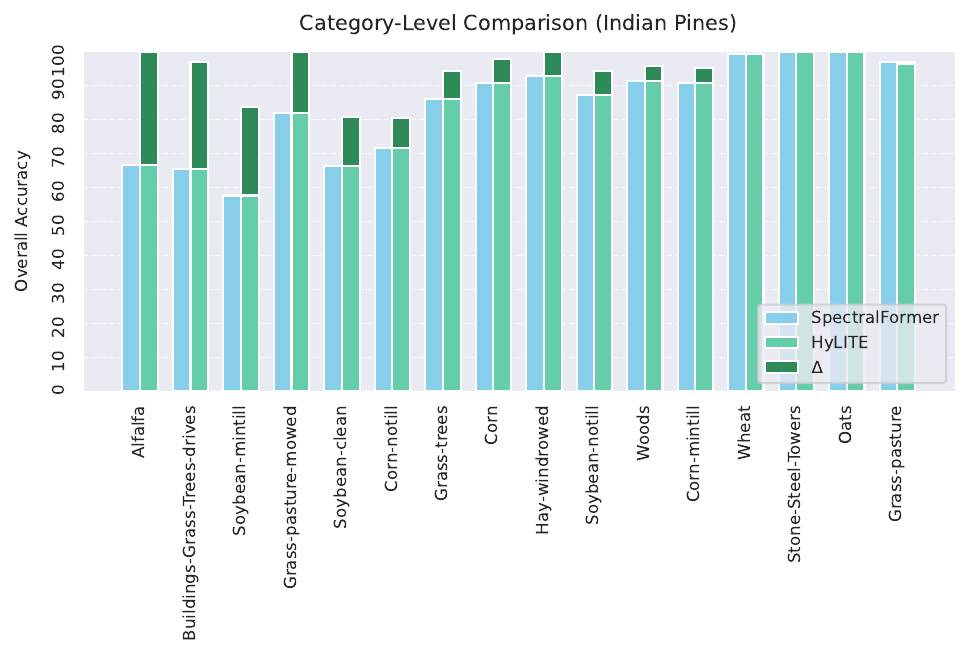}
\caption{Category-level comparison to the SpectralFormer on Indian Pines. The contribution of \system{} is generic, with fine-grained categories of 
`Alfalfa', `Buildings-Grass-Trees-drives', and `Soybean-mintill' receiving the highest benefits.}
% different \textbf{S-}oybeans, and \textbf{G-}rass receiving the highest benefit.}

%The improvement of \system{} is generic, with under-represented categories like Alfalfa with only $15$ training instances receiving the highest benefit from the locality.}
\label{fig:ip_classes}
\end{figure}

Here, we provide a category-level comparison with SpectralFormer on Indian Pines. Results are presented in Figure~\ref{fig:ip_classes}. The comparison results on Houston2013 and Pavia University datasets are presented in Section 3 of the supplementary material.

As can be seen, the contribution of \system{} is generic, as it improves over all (non-saturated) categories. The improvement is more pronounced for
easily misclassified classes by SpectralFormer, such as `Alfalfa', `Buildings-Grass-Trees-drives', and `Soybean-mintill'. This indicates that local, fine-grained details matter to distinguish such categories.
% different sub-categories of \textbf{S-}oybeans, and \textbf{G-}rass, which indicates that local, fine-grained details matter to distinguish such categories with only small visual differences. 

\vspace{-5mm}
\subsection{Architectural Analysis}

%Here, we analyse the contribution of different architectural choices. Results are presented in Table~\ref{tab:analysis}. Below, we detail each aspect separately. 

In this section, we provide further analysis to justify architectural choices in this paper (see Section~\ref{sec:method}). The results are presented in Table~\ref{tab:analysis}.

\begin{table}[t]
\centering
\resizebox{\columnwidth}{!}{
\begin{tabular}{@{}llllllllll@{}}
\toprule

 & \multicolumn{3}{c}{Indian Pines} & \multicolumn{3}{c}{Houston2013} & \multicolumn{3}{c}{Pavia University} \\ 
 \cmidrule(lr){2-10} 

 & OA & AA & Kappa & OA & AA & Kappa & OA & AA & Kappa \\ 
 \cmidrule(lr){2-4} \cmidrule(lr){5-7}  \cmidrule(lr){8-10}

%\multicolumn{10}{c}{\textit{Order of Attention}} \\ 
%\textit{} &   &  &  &  &  &  &  &  \\ 
\textit{Order of Attention} \\ 
\cdashlinelr{1-10}
Local-to-Spectral & 87.44  & 92.25  & 0.86  & 87.14  & 88.18  & 0.86  & 85.60  & 91.29  & 0.81 \\ 

\rowcolor{lightcyan} Spectral-to-Local & 89.80  & 94.69  & 0.88  & 88.49  & 89.74  & 0.87  & 91.28  & 92.25  & 0.88 \\ 

\midrule 
%\multicolumn{10}{c}{\textit{Token}} \\ 
\textit{Global Token} \\ 
\cdashlinelr{1-10}
Local-token & 83.54  & 92.61  & 0.81  & 85.85  & 87.26  & 0.85  & 91.05  & 92.16  & 0.88 \\ 

\rowcolor{lightcyan} Spectral-token & 89.80  & 94.69  & 0.88  & 88.49  & 89.74  & 0.87  & 91.28  & 92.25  & 0.88 \\ 

\midrule 

%\multicolumn{10}{c}{\textit{Fusion}} \\ 
\textit{Fusion} \\ 
\cdashlinelr{1-10}
Class-level & 81.05  & 89.86  & 0.79  & 86.05  & 87.83  & 0.85  & 87.41  & 88.20  & 0.83 \\ 

\rowcolor{lightcyan} Feature-level & 89.80  & 94.69  & 0.88  & 88.49  & 89.74  & 0.87  & 91.28  & 92.25  & 0.88 \\ 

\bottomrule
\end{tabular}}

\caption{Architectural Analysis. It is important to incorporate spectral attention prior to local, using a global spectral token as opposed to local, and combining representations at the feature-level rather than class-level.}
\label{tab:analysis}
\end{table}

\partitle{Order of Attention.} Firstly, we observe that the order of attentional blocks matters. Including spectral information prior to spatial information leads to better performance. This is expected, since in \hyper{} imaging, most of the information is present within the spectrum, and the input image may exhibit low spatial resolution. 

\partitle{Global Token.} Originally, the additional global classifier token for \system{} is spectral. Here, we test the performance with local tokens by transposing the dimensions and adding a global classifier token to the local dimension. We observe that pooling the representation from the spectral token matters, as opposed to local. This indicates that even though locality matters, spectral information carries more discriminative information for classification. 

%This is inline with our previous findings that local information is crucial to classify the \hyper{} images. 

\partitle{Fusion.} In \system{}, we combine spectral and local information directly within the block (\ie feature-level). We also experiment with late-fusion, where one model only includes spectral attention and the other only local attention, whose output is combined at the class-level. As is evident from Table~\ref{tab:analysis}, feature-level fusion outperforms the class-level counterpart by a large-margin, indicating the importance of spectral-to-local interactions for HSI. 

%In this work, we combine the local and spectral information directly at the feature-level, see the architecture in Figure~\ref{fig:architecture}. To understand the contribution of feature-fusion, we also train two separate transformers, where one only includes the local attention whereas the other only includes the spectral attention, and then combine their predictions at the classifier layer (class-level). As is evident, feature-level combination is much more superior to class-level fusion. This indicates that the interaction between local and spectral information is crucial to perform accurate classification. 

%The improvement is much more significant for categories with much fewer training examples: Alfalfa only has $15$ instances, confirming the efficacy of locality for \hyper{} scene understanding. 

\vspace{-4mm}
\section{Conclusion}

In this paper, we tackled \hyper{} image classification. Motivated by the limited local information in state-of-the-art \hyper{} image transformers, we incorporated the locality by attending to the local pixels as well as regularizing the local-to-global representations with our novel loss function. Evaluated on three well-established benchmarks, we observe that \system{} is highly accurate, as it improves state-of-the-art across all datasets and all metrics. Secondly, we observe that \system{} is highly efficient, as it learns much better from less number of examples. Finally, we highlight the importance of our architectural choices, for a principled inclusion of locality into \hyper{} transformers. We conclude that locality is crucial for \hyper{} image classification. 

%We hope our work inspires future research into better modelling the locality of the input imagery.  

%In this paper, we tackled \hyper{} image classification. Noticing the importance of locality, we proposed a locality-aware \hyper{} Transformer, \system{}, that models the local-spectral relationships via representation and regularization. We evaluated our model on three well-established benchmarks, shwocasing significant gain against all competing baselines across all metrics and settings. The contribution of \system{} is even more pronounced, when the training data is scarce. We conclude that locality is a strong and useful inductive bias for accurate and efficient \hyper{} image understanding. 

%We present an improved framework with additional spatial attention modules and extra global-local feature alignment loss objective based on the SepctralFormer~\citep{hong2021spectralformer} for HSIC. The spatial attention mechanism brings similarity information of an entire waveband between each neighbouring pixels and the global-local feature alignment technique aggregates local patch features to global representations. Thus, without pre-training requirement, our locality-aware approach gains better classification performance. As a future work, we will investigate the locality-aware strategy for self-supervised HSIC and HSI transfer learning such that it is more applicable for industries. We hope that our work inspires future methods in HSI tasks. 

%\mert{Compile all the materials in a compact manner here.}

\partitle{Acknowledgements.} This work was supported by the Dutch Science Foundation (NWO) under grant 700. Furthermore, we are grateful for the valuable and constructive feedback provided by our anonymous reviewers and the meta reviewer. Their input significantly contributed to the enhancement of our manuscript.
\bibliography{egbib}
\end{document}